\documentclass[12pt,onecolumn,letterpaper]{article}

\usepackage{cvpr}
\usepackage{times}
\usepackage{epsfig}
\usepackage{graphicx}
\usepackage{amsmath}
\usepackage{amssymb}
\usepackage{booktabs}

\usepackage[pagebackref=true,breaklinks=true,letterpaper=true,colorlinks,bookmarks=false]{hyperref}

\cvprfinalcopy

\usepackage{graphicx}
\usepackage{times}
\usepackage{graphicx}
\usepackage{amsmath}
\usepackage{amssymb}
\usepackage{url}
\usepackage{times}
\usepackage{epsfig}
\usepackage{multirow}
\usepackage{algorithm}
\usepackage{algorithmic}

\makeatletter

\newcommand{\Rmnum}[1]{\expandafter\@slowromancap\romannumeral#1@}
\makeatother

\def\Mat#1{{\boldsymbol{#1}}}

\begin{document}
%
\title{SlideNet: Fast and Accurate Slide Quality Assessment Based on Deep Neural Networks}




\author{Teng Zhang$^1$, Johanna Carvajal$^1$,
Daniel F. Smith$^1$,
Kun Zhao$^1$, \\
Arnold Wiliem$^1$, 
Peter Hobson$^2$,
Anthony Jennings$^2$,
and Brian C. Lovell$^1$ \\
$^1$School of ITEE, The University of Queensland, Australia \\
$^2$Sullivan Nicolaides Pathology, Australia
}


\maketitle

\begin{abstract}
This work tackles the automatic fine-grained slide quality assessment problem for digitized direct smears test using the Gram staining protocol. Automatic quality assessment can provide useful information for the pathologists and the whole digital pathology workflow. For instance, if the system found a slide to have a low staining quality, it could send a request to the automatic slide preparation system to remake the slide. If the system detects severe damage in the slides, it could notify the experts that manual microscope reading may be required. In order to address the quality assessment problem, we propose a deep neural network based framework to automatically assess the slide quality in a semantic way. Specifically, the first step of our framework is to perform dense fine-grained region classification on the whole slide and calculate the region distribution histogram. Next, our framework will generate assessments of the slide quality from various perspectives: staining quality, information density, damage level and which regions are more valuable for subsequent high-magnification analysis. To make the information more accessible, we present our results in the form of a heat map and text summaries. Additionally, in order to stimulate research in this direction, we propose a novel dataset for slide quality assessment. Experiments show that the proposed framework outperforms recent related works. 

\end{abstract}


%

\section{Introduction}

Microscopy image analysis plays a crucial role in diagnosing various types of infection and disease. The distribution of bacteria and cells can provide valuable information for pathologists. Nonetheless, this important process is a very time-consuming task which can only be performed by experienced scientists and pathologists. To ease the workflow for experts and solve the labor intensity issue, several simple Computer Aided Diagnosis (CAD) systems have been developed in recent years~\cite{tadrous2010mycobacteria}\cite{ bibbo1991}\cite{lewis2012}\cite{Johanna2017}\cite{gurcan2009}. For example,~\cite{tadrous2010mycobacteria} only considered rejecting easy negative samples. In~\cite{bibbo1991}\cite{lewis2012}, the authors performed basic cell counting tasks on manually pre-selected regions. More recently, coarse-grain region classification has been investigated~\cite{Johanna2017}\cite{ gurcan2009}. Although progress has been made, most existing works do not consider high-level tasks such as slide quality assessment and they generally rely on traditional machine learning methods such as support vector machine (SVM) with hand-crafted features. 

As mentioned above, the fine-grained slide assessment problem remains unsolved in the literature. However, it is extremely important and useful for an automatic slide scan and analysis system as it can report a variety of issues from the assessment. It can also assist pathology labs to reduce mistakes, provide data for quality metrics, and save effort. 

\textbf{Problem Definition:} We define this novel problem as follows: Given a sample slide $\Mat{T}$, we need a fast and accurate system $\Mat{\Phi}$ that produces various useful outputs $\Mat{O} = \Mat{\Phi(T)}$. The outputs $\Mat{O}$ should include an information density map, and assessment text summaries such as the staining quality, information density, and the damage level. Note that the solution to this problem needs to be efficient as the digitized slide image is usually very large.

\begin{figure}[t]
  \centering
  \includegraphics[width=1 \linewidth]{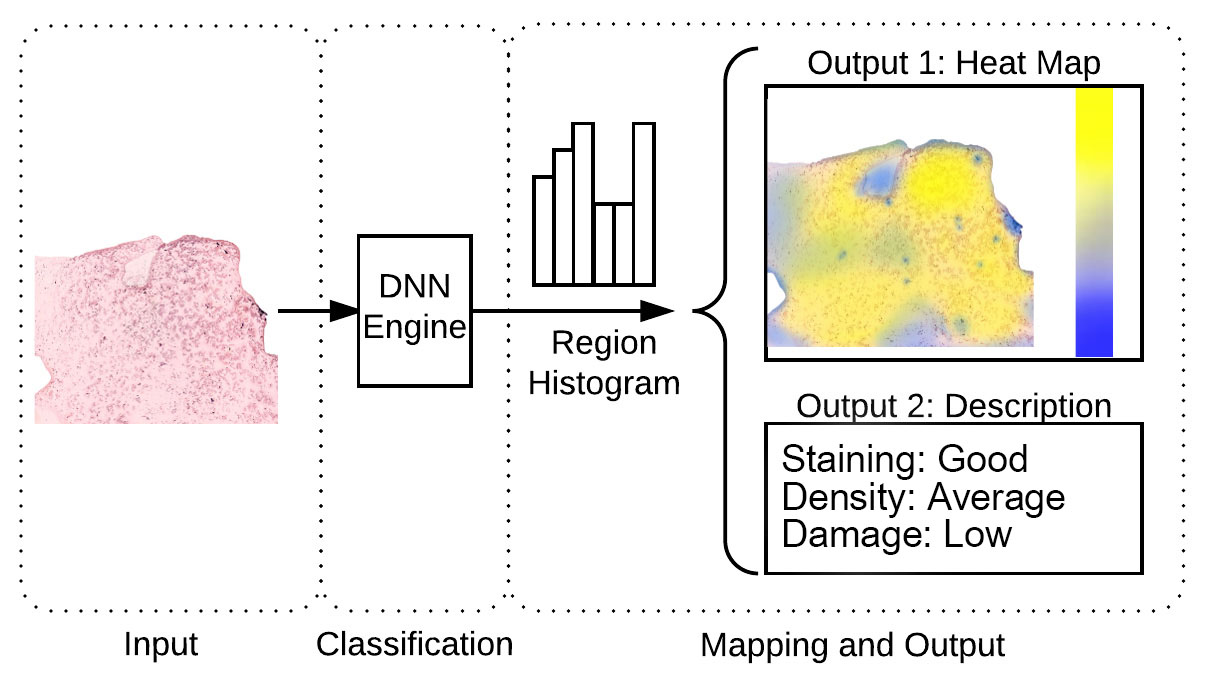}
  \caption{Our proposed SlideNet framework for assessing the slide quality and calculating the content distribution. Firstly, our DNN engine performs a fast dense region classification on the entire slide. Then, the heat map and text summaries are generated automatically based on the classification results.}
 \label{figure_framework}
\end{figure}

In this work, we aim to solve this problem by performing fast and accurate dense region classification on the entire slide in a fine-grained style. This provides the content distribution of the target slide to draw conclusions and assessments. To perform the dense region classification task, we noticed that traditional methods using SVM with hand-crafted features are not accurate compared to deep learning methods. However, we can not simply turn to deep learning as it tends to be slower compared to traditional methods and it also requires significant computing power which can be absent in a microscope related scenario. Luckily, recent progress~\cite{mobilenet2017}\cite{Squeezenet}\cite{shufflenet2017} in small deep neural networks makes it possible to run DNN using less computing power and storage while maintaining reasonable accuracy. By employing small DNNs, we are able to construct our slide assessment framework, SlideNet, that can produce the required outputs $\Mat{O}$ efficiently, as shown in Figure 1. Associated with this novel problem, a new fine-grained Gram stain slide analysis dataset is also constructed. 

\textbf{Contributions}: We list our contributions as follows:
\begin{enumerate}
\item To the best of our knowledge, this is the first work investigating automatic assessment of slide quality and content distribution in a fine-grained perspective for Gram stain slides. We propose a DNN based framework to address this novel problem.
\item We propose a new fine-grained Gram stain slide sample dataset to fill the research gap and evaluate our proposed framework. This dataset will be made publicly available.
\item  We achieved fast speed with low memory cost by employing a tiny DNN, namely MobileNet, as our dense classification engine and leading to much higher accuracy compared to recent related works in the region selection task.
\end{enumerate}
This paper is continued as follows: Section 2 introduces
the related works and the proposed dataset is described
in Section 3. We then introduce the proposed framework and baseline methods in Section 4. Experimental results are provided and discussed in Section 5. Finally, we conclude this paper in Section 6.

\section{Related Works} 

In this section, three related topics are discussed: (1) Automation in microscopic medical image analysis; (2) Small neural networks; (3) The limited labelled data problem.

\subsection{Automation in microscopic medical image analysis}

As mentioned in the previous section, we evaluate our framework on the Gram stain test. We briefly review recent progress mainly in Gram stain tests and will also include other related microscope based slide tests. 

In the scope of Gram stain test, most related works~\cite{Chayadevi2012}\cite{crossman2015}\cite{ Hiremath2010} only aim to detect and count the existence of different cells and bacteria. Despite the fact that progress has been made, these works assume that the areas of interest have been pre-selected by experts. As such, the fast automatic area of interest selection~\cite{Johanna2017} in low magnitude objectives is becoming an important step to have a fully automated CAD system for this application. The reason is that the final analysis is done by using high-powered objectives (e.g. 100x) which involves extremely long scanning times, huge image storage, and long processing times. So in the real scenario, it is a common strategy to choose a subset of good regions in low-mag to zoom into for further high-mag analysis. 

In a more general scope, an important category of microscopic image analysis works is the design of the feature descriptor. For example,~\cite{Kato2015}\cite{Sanroma2014} use Histogram of Gradients (HOG)~\cite{Dalal2005} as the feature descriptor to perform kidney slice analysis and brain image analysis, respectively. Local binary pattern (LBP) is also selected as the descriptor in several related works such as oral cancer identification~\cite{Krishnan2012}, breast tissue classification~\cite{SunhuaWan2017}, and Human Embryo classification~\cite{LiangXu2014}. Also, 2D-DCT has also been employed in identifying organelles from fluorescence microscopy
images of HeLa cells~\cite{XiaopengHong2016} and detection of Acute Lymphoblastic Leukemia~\cite{Mishra2017}.

In the category of low-level digital image processing, there have been several simple coarse-grain slide quality assessment works~\cite{Ameisen2013}\cite{Avanaki2016}. However, the disadvantages of these works are quite obvious: 1. They are based on digital image processing instead of computer vision and machine learning so the content analysis of the slide is missing and the results are not satisfying. 2. More importantly, they focused more on the image (sharpness, saturation and brightness) rather than the slide itself (damage level, staining quality and tissue ratio).

More recently, deep learning methods are also being investigated in several microscopic medical applications~\cite{Ravishankar2016}\cite{DinggangShen2016}. However, these works address different problems thus their frameworks are not suitable for our proposed tasks. Perhaps the most similar work to us is~\cite{Johanna2017} where the authors use SVM to select the regions of interest in a coarse-grain style using hand-crafted feature descriptors. In contrast, we differ significantly: 1. We propose a novel fine-grained slide quality assessment problem with a new larger, improved, and more difficult dataset. 2. We propose a novel DNN based framework that is 20x faster in training, 4x faster in testing and more accurate compared to~\cite{Johanna2017} in the region classification task.

\subsection{Small neural networks}

Deep neural networks often come with millions of parameters which requires weeks of training and a huge amount of training data. However, in many real-world applications, the computing power and on-chip memory are limited in most embedded devices. Also, large deep learning models usually are slower compared to traditional machine learning techniques such as SVM. 

Recently, researchers have been working on more efficient deep neural network models such as MobileNet~\cite{mobilenet2017}, SqueezeNet\cite{Squeezenet} and ShuffleNet\cite{shufflenet2017}. These works have successfully led to smaller model sizes and less computation while maintaining reasonable accuracy. Among these, MobileNet has been tested thoroughly and multiple pre-trained MobileNet models have been released. A more in-depth discussion on small networks can be found in~\cite{keynote_smallNN}.

With all of these considerations, we employ a small DNN, namely, MobileNet, in our framework for the best trade-off between accuracy and speed. 

\subsection{The limited labelled data problem}

DNN based methods usually need a huge amount of labelled data which may not be available. Recently, researchers have proposed several ways to address this problem. For example, Minimalistic Directed Generative Networks~\cite{simpltron2015} has shown potential in training with few labels. Luo~\cite{Luo2017} proposed a hierarchical active learning framework that divides the whole population into smaller groups, then a more refined model can be gradually learned from the groups and soft labels.

However, a more common and realistic strategy is to fine-tune a pre-trained model on the new dataset. The pre-trained model is usually trained on large dataset such as ImageNet~\cite{ILSVRC15}. We follow this path as MobileNet has provided various pre-trained models. 

\section{A Novel Fine-Grained Gram Stain Dataset}

Gram stain is a common staining technique used in microbiology to distinguish bacterial species into two broad groups: Gram-positive and Gram-negative~\cite{Singleton2007}. Christian Gram developed this technique in 1883~\cite{Bartholomew1952}. With this staining technique, the bacteria will be stained either pink (Gram-negative) or violet (Gram-positive) depending on the cell wall properties. By observing the distribution of different positive or negative bacteria and cells, doctors can draw some important conclusions in diagnosing infections and diseases.

One publicly available dataset with Gram stain images
focuses on the classification of micro-organisms within the
image slides~\cite{crossman2015}. The dataset used in~\cite{crossman2015} contains 150 colour
images with positive and negative examples of leukocyte
and epithelial cells. Another dataset presented in~\cite{Chayadevi2012} contains
320 stained slide images not only using the Gram stain
technique but also other different staining techniques. The
problem addressed in~\cite{Chayadevi2012} is the automatic detection of microbes and the extraction of bacterial clusters.

There is only one dataset~\cite{Johanna2017} that can be used for the task of selecting candidate areas for subsequent high-magnification
analysis. This dataset consists of eight Gram stain images. They used a Zeiss Imager Z2 microscope and the images are captured using a PixeLINK PL-B623CF camera. However, this dataset is relatively small and only has 3 coarse labels. As such, it does not support fine-grained region analysis and therefore cannot be used for the proposed tasks of slide quality assessment and information density map.

\begin{figure}[htb]
  \centering
  \includegraphics[width=0.75 \linewidth]{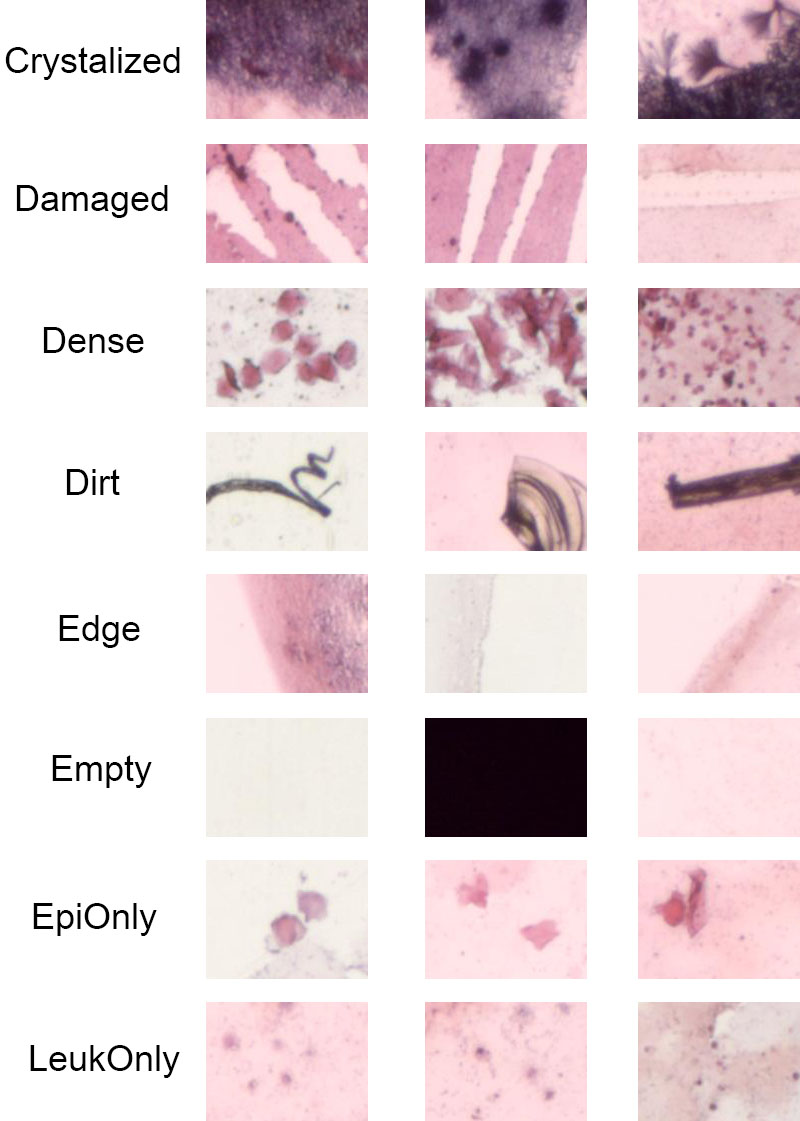}
  \caption{The selected samples of the proposed dataset. Best viewed in color.}
 \label{figure_dataset}
\end{figure}

To study our proposed novel tasks, we collected a medium size benchmarking dataset during late 2017. We used a Zeiss Imager Z2 microscope which is the same as~\cite{Johanna2017} but we employed a much better camera that has a 12 mega-pixel sensor to capture the images. With the help of scientists and pathologists, we managed to collect 4,000 patch images and label them in 8 fine-grained classes: Crystalized, Damaged, Dense, Dirt, Edge, Empty, EpiOnly and LeukOnly. Among these, 3000 images are used for training and 1000 images for testing. Additionally, we provide 6 whole slide images for task 2. The dataset will be available for download at \url{http://www.itee.uq.edu.au/sas/digital-pathology-datasets}. The selected samples are provided in Figure~\ref{figure_dataset}. We explain the meaning of the labels below. 

\textbf{1. Crystalized} is a region where the staining dye became crystalized. It is typically caused by bad staining technique or aged staining chemicals and it greatly reduces the information level of that area.

\textbf{2. Damaged} is a region which became scratched and damaged during storage, transportation or observation. It is typically caused by careless human handling and it can erase some information from that area. 

\textbf{3. Dense} is a region filled with many Epithelials and Leukocytes. It represents a good region to observe later and it has the highest level of information density.

\textbf{4. Dirt} is a region covered with dirt, hair or dust. They may contain useful information but it may also cause problems for the automatic focus algorithms.

\textbf{5. Edge} is a region located at the edge of the tissue area. The information density level of this type of region is average or above-average in some cases.

\textbf{6. Empty} is a region located at the empty part of the slide. It is the worst region to select for further analysis and its information density level is the lowest of all labels.

\textbf{7. EpiOnly} is a region which only has three or fewer notable Epithelials. It is rated as a good selection in general and its information density level is above average.

\textbf{8. LeukOnly} is a region which only has a few or less notable Leukocytes. It is rated as a good selection in general and its information density level is above average.

Note that our proposed region analysis framework can be adapted and work for different types of slides. As we have the Gram stain slides dataset ready and labelled, our proposed framework is only tested on Gram Stain slides at this stage.

\section{Proposed DNN based Framework}

As shown in Figure 1, the proposed framework has two stages: dense region classification and the follow on slide quality assessment. Details are provided below.

\subsection{Region Classification Engine}

Fast and accurate fine-grained region classification for the entire slide is the foundation of our framework. The problem is challenging due to the large intra-class variance as well as the inter-class similarity in some cases. Traditional hand-crafted features will most likely fail. Luckily, with the rapid development of CNNs, image classification has taken a huge step forward. As mentioned in the related works, we built our framework on MobileNet~\cite{mobilenet2017} for the best trade-off between speed and accuracy. We now briefly introduce the key component of MobileNet.

A standard convolution operation takes the input $D_F\times D_F\times M$ feature map $\Mat{F}$ and produces the output $D_G\times D_G\times N$ feature map $\Mat{G}$ where $D_F$ is the spatial width and height of a square input feature map, $M$ is the number of input channels (input depth), $D_G$ is the spatial width and height of a square output feature map, and $N$ is the number of output channels (output depth). 

However, MobileNet is based on depth-wise separable convolution which is a form of factorized convolutions that factorize a standard convolution into a depth-wise convolution and a 1$\times$1 convolution called a point-wise convolution~\cite{mobilenet2017}. In other words, the convolution layer in MobileNet is split into two layers: depth-wise layer and point-wise layer.

So the computation cost of the standard convolution: 
\begin{equation}
D_K \cdot D_K \cdot M \cdot N \cdot D_F \cdot D_F
\end{equation}
\noindent
will be reduced to:
\begin{equation}
D_K \cdot D_K \cdot M \cdot D_F \cdot D_F + M \cdot N \cdot D_F \cdot D_F
\end{equation}
\noindent
which is the sum of the depth-wise and $1\times 1$ point-wise convolutions and $D_K$ is the spatial dimension of the kernel.

Thus, the model size and computation complexity are both greatly reduced. Interested readers can refer to~\cite{mobilenet2017} for a full treatment of MobileNet.

From various applications in the literature~\cite{Parkhi15}\cite{resNet16}\cite{Zhang2018}, we know that a well trained deep CNN feature descriptor is much more powerful than traditional hand-crafted features. However, as discussed in Section 2, a major challenge in the medical image analysis community is the limited data problem. Fine-tuning a pre-trained model can be one of the solutions. In our framework, we fine-tune the last layer of the pre-trained MobileNet model on our proposed dataset. 

\subsection{Mapping and Output}

Based on the efficient MobileNet classification engine, we can perform dense region classification on the entire slide and calculate the distribution of different region types. After that, it is possible to generate an information density mask for the original slide which we call a heat map. Directed by pathologists, we defined the region information density value $I(r)$ of region $r$ based on (3) and (4). The coloring for the region $r$ is based on the calculated information density value $I(r)$. We choose yellow and blue as the indicators where yellow means more information and blue means less information. We have

\begin{equation}
I(r)=\lambda_1\cdot\mathcal{S}(l_1(r))+\lambda_2\cdot\mathcal{S}(l_2(r))
\end{equation}
\noindent where $\mathcal{S}(\cdot) \in [0, 255]$ is the intensity score function for a certain label. $l_1(r)$ and $l_2(r)$ are the top two predicted labels respectively. $\lambda_1$ and $\lambda_2$ are set to 0.8 and 0.2 empirically. We discretely get the value of $\mathcal{S}(\cdot)$ from a simple parabolic:
\begin{equation}
\mathcal{S}(l) = -x^2+\lambda_3, 
\begin{cases}
x = 1, &\text{if $l$ = Dense} \\
x = 8, &\text{if $l$ = EpiOnly or LeukOnly} \\ 
x = 13, &\text{if $l$ = Edge or Damaged}  \\
x = 15, &\text{if $l$ = Crystalized or Dirt}\\
x = 16, &\text{if $l$ = Empty}
\end{cases}
\end{equation}
\noindent where $\lambda_3$ is set to 256 to match the scope of $\mathcal{S}(\cdot)$ and $x$ is empirically selected on the curve to represent different labels.

Another output based on the region distribution is the slide quality assessment text description based on simple comparison of region distribution ratio. Keywords (Good/Average/Low) are mapped from empirically selected thresholds. An example output of our proposed SlideNet framework is given in Figure 3. We have selected a small region to show the difference in color saturation.

\begin{figure}[htb]
  \centering
  \includegraphics[width=0.9 \linewidth]{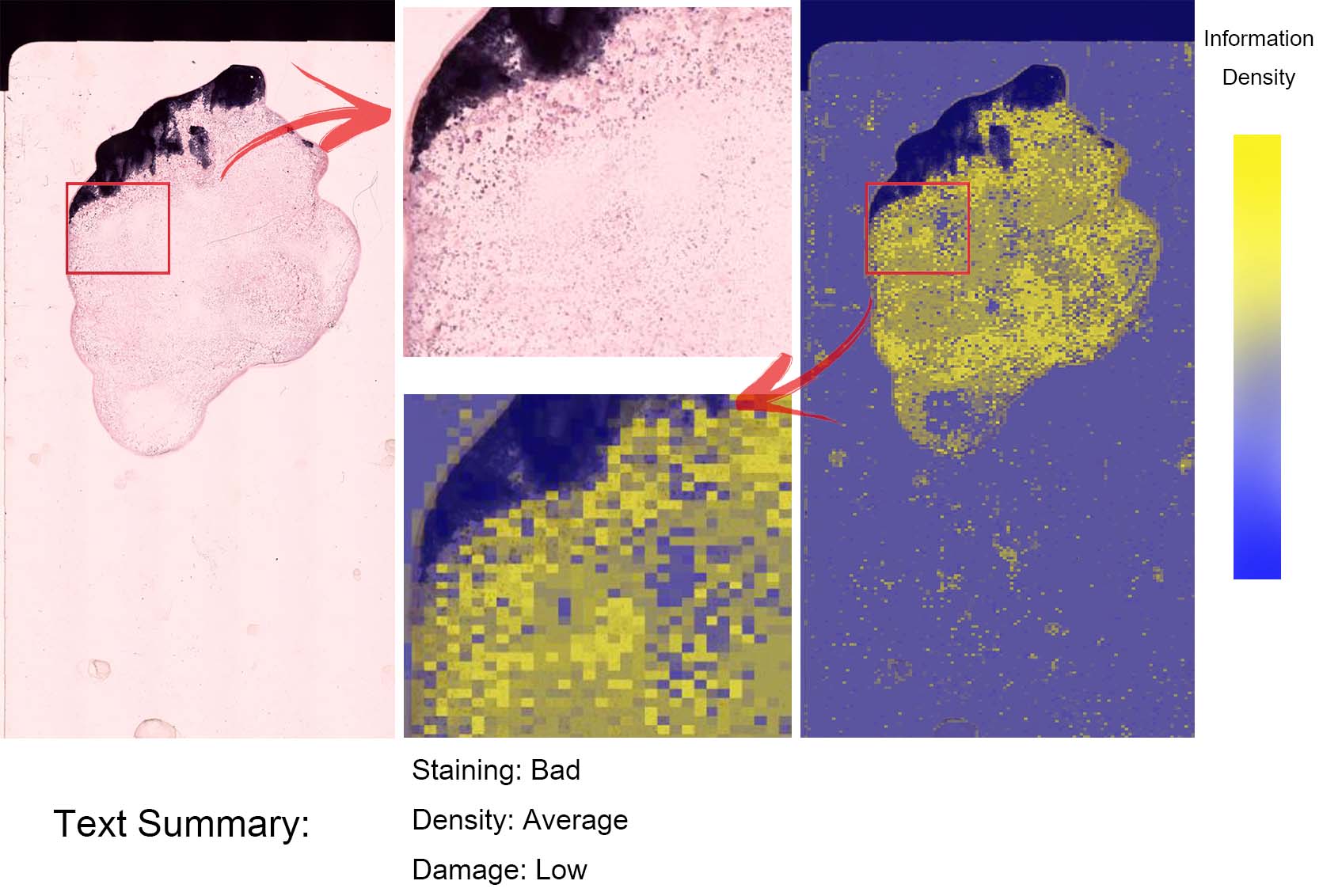}
  \caption{The example output of SlideNet: The slide will be parsed and displayed in an information density mask (heat map). Also, text summaries will be generated automatically based on region distribution ratios.}
 \label{figure_task2}
\end{figure}

\section{Experiments}

\begin{figure*}[htb]
  \centering\includegraphics[width=0.95 \linewidth]{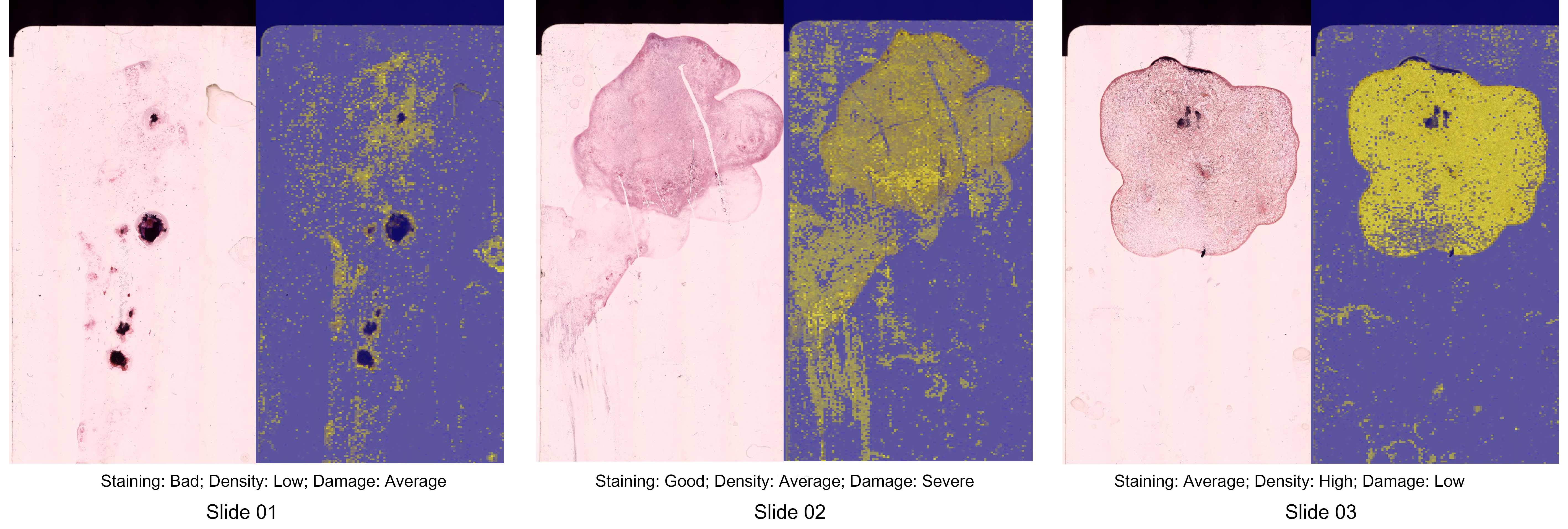}\par
  \caption{Visualization of selected results. These slides are also included in our dataset. Best viewed in color.}
  \label{figure_visualization}
\end{figure*}%

All evaluations were done using NVidia K40c GPU
with the TensorFlow framework in python. In addition, Adam optimizer~\cite{Adam2014} was used with a training batch size of 100. The learning rate is set to 0.01.

\subsection{Task 1:Region Classification}
Before we dive into our theme, we first show the dense region classification task performance. As shown in Table~\Rmnum{1}, colorLBP+SVM~\cite{Johanna2017} can run in 50FPS and has an accuracy of 51.21\% while HOG+SVM~\cite{Dalal2005} achieves slightly worse performance (45.3\%) with 5 less fps. On the contrary, SlideNet based on MobileNet engine with different network sizes (128 or 224)~\cite{mobilenet2017} can achieve much higher accuracy (82.8\% or 86.8\%) and the speed can be accelerated using GPU. 

\begin{table}[htbp]
\centering
\caption{Fine-Grained Slide Region Classification Results.}
\begin{tabular}{l |c|c}
\hline
Classification Engine &Speed &Accuracy \\ \hline
HOG+SVM~\cite{Dalal2005}    &45FPS(CPU)    &45.3\% \\ \hline
colorLBP+SVM~\cite{Johanna2017}      &50FPS(CPU)  &51.2\%  \\ \hline
SlideNet-128       &14FPS(CPU), \textbf{200FPS}(GPU)  &82.8\% \\ \hline
SlideNet-224       &10FPS(CPU), 170FPS(GPU)  &\textbf{86.8\%} \\ \hline
\end{tabular}
\label{table_task_region_classify}
\end{table}

We can see that those baseline methods with hand-crafted features perform much worse compared to the proposed deep CNN based SlideNet.

\subsection{Task 2: Density Heat Map and Quality Assessment}

As mentioned above, the main goal of this paper is to assess the quality of the slide and transfer the slide related knowledge of scientists and pathologists to the proposed framework, SlideNet. The evaluation for these functions are provided below.

\textbf{Heat Map and Assessment Visualization:} From the classification results, we will have the distribution of different regions for a certain slide. After that, the map and descriptions are generated following Section 4-B. The complete assessment and visualization of selected slides are provided in Figure 4.

\textbf{Subjective Evaluation:} To evaluate the SlideNet's predictions, we invited three experts to comment on the original slide images at the same time to measure the correlation between SlideNet and human experts, and estimate the accuracy. The opinions of the experts are in agreement from voting.

Since the ranking metric is the same (either Good/High/Severe, Average or Bad/Low) for both SlideNet or human experts, we can simply count a same ranking as 1 point and a different ranking as 0 point. Based on the 6 whole slide images, we can calculate the similarity between SlideNet and the human experts.
\begin{small}
\begin{table}[htbp]
\centering
\caption{Subjective Comparison: SlideNet versus Experts}
\begin{tabular}{l |p{12mm}| p{10mm} | p{10mm}  |  l c}
\hline
Slide ID  &Attribute &SlideNet  & Experts &Accuracy\\ \hline
01  &Staining: Density: Damage: & Bad Low Average & Bad Low Low  & 0.66\\ \hline
02  &Staining: Density: Damage: & Good Average Severe & Good Average Severe &1.0\\ \hline
03  &Staining: Density: Damage: & Average High Low  & Good High Low &0.66\\ \hline
04  &Staining: Density: Damage: & Good Average  Low & Good High Low &0.66\\ \hline
05  &Staining: Density: Damage: & Good  High Low &Good High Low  & 1.0\\ \hline
06  &Staining: Density: Damage: &  Average Average Low& Average High Low& 0.66\\ \hline
Overall      & - & -  &-  & \textbf{0.778}\\ \hline                 
\end{tabular}
\label{table_task_quality}
\end{table}
\end{small}

From Table~\Rmnum{2}, we can see that SlideNet can give reasonably accurate predictions (77.8\%) with limited training data and time. Note that these are early results and there will be more challenges in the future work. However, we argue this is one key milestone in the full automation of Microscopic medical image analysis.

\section{Conclusion}

In this paper, aiming to direct the microscope's actions and assess the slide quality without supervision, we proposed a fast framework named SlideNet and a fine-grained region classification dataset together with several whole slide images for Gram stain slides. We showed that it is possible to generate meaningful slide assessments and provide data for quality metrics. Also, we are able to tell the scanning system which regions are better regions to scan using higher magnification. In the future, we will expand the scale of the dataset to include other types of slides, and we will work on high magnitude cell and bacteria detection and classification. 

\section{Acknowledgements}
This work has been funded by Sullivan Nicolaides
Pathology, Australia and the Australian Research Council
(ARC) Linkage Projects Grant LP160101797. Arnold Wiliem
is funded by the Advance Queensland Early-Career Research
Fellowship.






%
{\small
\bibliographystyle{IEEEtran}
\bibliography{egbib}
}

\end{document}